\pdfoutput=1
\documentclass[11pt]{article}

\usepackage[T1]{fontenc}
\usepackage{arabtex}
\usepackage[]{acl}
\usepackage{multirow}
\usepackage{graphicx}
\usepackage{array}
\usepackage{times}
\usepackage{latexsym}
\usepackage[utf8]{inputenc}
\usepackage{utf8}
\usepackage{xspace}
\usepackage{tabularray}
\usepackage{geometry}
\usepackage{microtype}
\usepackage{makecell}

\setcode{utf8}
\newcommand{\Ar}[1]{{\scriptsize\<#1>}}

\title{Sina at FigNews 2024:\\ Multilingual Datasets Annotated with Bias and Propaganda}

\author{
  \begin{minipage}[t]{0.30\textwidth}
    \centering
    Lina Duaibes \\
    \textnormal{Birzeit University} \\
    \textnormal{1205358@student.birzeit.edu}
  \end{minipage}
  \hfill
  \begin{minipage}[t]{0.30\textwidth}
    \centering
    Areej Jaber \\
    \textnormal{Palestine Technical University-Khadoorie} \\
    \textnormal{a.jabir@ptuk.edu.ps}
  \end{minipage}
  \hfill
  \begin{minipage}[t]{0.30\textwidth}
    \centering
    Mustafa Jarrar \\
    \textnormal{Birzeit University} \\
    \textnormal{mjarrar@birzeit.edu}
  \end{minipage}
  \\[4em] % Increase the vertical space between the rows
  \begin{minipage}[t]{0.45\textwidth}
    \centering
    \textbf{Ahmad Qadi} \\
    7amleh Center \\
    \textnormal{ahmad@7amleh.org}
  \end{minipage}
  \hfill
  \begin{minipage}[t]{0.45\textwidth}
    \centering
    \textbf{Mais Qandeel} \\
    ÖREBRO University \\
    \textnormal{Mais.Qandeel@oru.se}
  \end{minipage}
}

\begin{document}

  \maketitle

\begin{abstract}
The proliferation of bias and propaganda on social media is an increasingly significant concern, leading to the development of techniques for automatic detection. This article presents a multilingual corpus of $12,000$ Facebook posts fully annotated for bias and propaganda. The corpus was created as part of the FigNews $2024$ Shared Task on News Media Narratives for framing the Israeli War on Gaza. It covers various events during the War from October $7$, $2023$ to January $31$, $2024$. The corpus comprises $12,000$ posts in five languages (Arabic, Hebrew, English, French, and Hindi), with $2,400$ posts for each language. The annotation process involved $10$ graduate students specializing in Law. The Inter-Annotator Agreement (IAA) was used to evaluate the annotations of the corpus, with an average IAA of $80.8\%$ for bias and $70.15\%$ for propaganda annotations. Our team was ranked among the best-performing teams in both Bias and Propaganda subtasks. The corpus is open-source and available at {\small \url{https://sina.birzeit.edu/fada}}
\end{abstract}

\section{Introduction}

Since October $7$, social media has been flooded with posts, articles, images, and videos related to the Israeli War on Gaza. Such posts are often divided by hate, bias, and fake news either in favor of or against one of the parties or by remaining neutral, see e.g., 
"Framing the Israeli War on Gaza" is a shared task on news media narratives \cite{zaghouani2024arabicnlp}, which is part of the $2$\textsuperscript{nd} ArabicNLP conference. The task aims to create a multilingual corpus that unravels the layers of bias and propaganda within news articles in various languages.

Such shared tasks and datathons are crucial in the NLP community to foster collaboration and advance research in specific areas. Previous efforts, such as 
SemEval-$2020$ Task $11$ \cite{martino2020semeval} and TSHP-$17$ \cite{rashkin2017truth} have provided valuable resources for propaganda detection in news articles. The dual focus of FigNews on bias and propaganda is a novel approach that addresses the evolving nature of misinformation on social media platforms.
The detection of propaganda on social media is crucial \cite{DH21}, as it can polarize public sentiment, foster violent extremism and hate speech, and eventually erode democracies and diminish trust in democratic procedures \cite{ARJ17}. Notably, only a few corpora have been recently built to address these issues. Recent work by \citep{HJKN23} involved establishing a Hebrew dataset comprising $15,881$ tweets for detecting offensive language. This dataset was manually annotated with four labels: hate, abusive, violence, and pornographic. Their work focused on detecting hate speech in Hebrew tweets and implemented in \textit{SinaTools} \cite{HJK24}. Additionally, the WojoodNER Shared Task 2024 offered a new NER dataset related to the Israeli War on Gaza called  \textit{Wojood\textsuperscript{Gaza}} \cite{JHKTEA24}.
Other notable works include TSHP-$17$ \citep{rashkin2017truth}, QProp \citep{barron2019proppy}, and PTC \citep{da2019fine}. TSHP-$17$ and QProp are document-level corpora, while PTC is a sentence-level corpus. While SemEval-$2020$ \citep{martino2020semeval} Task $11$ is similar to FigNews \citep{zaghouani2024arabicnlp} in its objective, they differ in their data sources and focus areas.

This paper describes our participation in the FigNews. Our contributions are:
\begin{itemize}
    \item Annotated Corpus ($12K$ FB posts) for bias and propaganda, in $5$ languages.
  \item Annotation guidelines ensuring consistency and accuracy.
    \end{itemize}

\textbf{Remark}: The corpus presented in this article does not cover the genocide, ethnic cleansing, or starvation events as they mostly happened after collecting the corpus.

The article is organized as follows: Section \ref{sec_methodology} describes the methodology, \ref{sec_team} presents our team composition and training; 
Section \ref{sec3} presents our participation and results; Section \ref{sec_discussion} analyzes some errors, and Section \ref{sec_conclusion} concludes the paper.

\section{Annotation Methodology}
\label{sec_methodology}

The objective of the task is to address the complex landscape of social media discourse related to the Israeli War on Gaza $2023$-$2024$.
The task organizers provided participants with $15$k posts from verified Facebook accounts, selected between October $6$, $2023$, and January $31$, $2024$, using "Gaza" as a query keyword across $5$ languages: Arabic, Hebrew, English, French, and Hindi. The dataset consists of $15$ batches, each containing $1000$ posts. 

\begin{table*}[ht]
\small
\centering
\begin{tabular}{cc|cccc|cccc|}
\cline{3-10}
 &  & \multicolumn{4}{c|}{Biased} & \multicolumn{4}{c|}{Propaganda} \\ \hline
\multicolumn{2}{|c|}{} & \multicolumn{2}{c|}{Cohen’s kappa} & \multicolumn{2}{c|}{F1\_score\_weighted} & \multicolumn{2}{c|}{Cohen’s kappa} & \multicolumn{2}{c|}{{\color[HTML]{4C4C4C} F1\_score\_weighted}} \\ \cline{3-10} 
\multicolumn{2}{|c|}{\multirow{-2}{*}{Annotators’ pair}} & \multicolumn{1}{c|}{All} & \multicolumn{1}{c|}{Binary} & \multicolumn{1}{c|}{All} & Binary & \multicolumn{1}{c|}{All} & \multicolumn{1}{c|}{Binary} & \multicolumn{1}{c|}{All} & Binary \\ \hline
\multicolumn{1}{|c|}{} & 6 & \multicolumn{1}{c|}{0.57} & \multicolumn{1}{c|}{0.57} & \multicolumn{1}{c|}{0.79} & 0.79 & \multicolumn{1}{c|}{0.76} & \multicolumn{1}{c|}{0.85} & \multicolumn{1}{c|}{0.85} & 0.98 \\ \cline{2-10} 
\multicolumn{1}{|c|}{} & 4 & \multicolumn{1}{c|}{0.76} & \multicolumn{1}{c|}{0.77} & \multicolumn{1}{c|}{0.53} & 0.56 & \multicolumn{1}{c|}{0.33} & \multicolumn{1}{c|}{0.25} & \multicolumn{1}{c|}{0.58} & 0.93 \\ \cline{2-10} 
\multicolumn{1}{|c|}{} & 8 & \multicolumn{1}{c|}{0.29} & \multicolumn{1}{c|}{0.28} & \multicolumn{1}{c|}{0.64} & 0.65 & \multicolumn{1}{c|}{0.11} & \multicolumn{1}{c|}{1} & \multicolumn{1}{c|}{0.28} & 1 \\ \cline{2-10} 
\multicolumn{1}{|c|}{\multirow{-4}{*}{1}} & 2 & \multicolumn{1}{c|}{0.64} & \multicolumn{1}{c|}{0.64} & \multicolumn{1}{c|}{0.82} & 0.85 & \multicolumn{1}{c|}{0.72} & \multicolumn{1}{c|}{0.62} & \multicolumn{1}{c|}{0.8} & 0.91 \\ \hline
\multicolumn{1}{|c|}{} & 10 & \multicolumn{1}{c|}{0.8} & \multicolumn{1}{c|}{0.78} & \multicolumn{1}{c|}{0.86} & 0.9 & \multicolumn{1}{c|}{0.67} & \multicolumn{1}{c|}{1} & \multicolumn{1}{c|}{0.77} & 1 \\ \cline{2-10} 
\multicolumn{1}{|c|}{} & 4 & \multicolumn{1}{c|}{0.51} & \multicolumn{1}{c|}{0.55} & \multicolumn{1}{c|}{0.75} & 0.79 & \multicolumn{1}{c|}{0.81} & \multicolumn{1}{c|}{1} & \multicolumn{1}{c|}{0.87} & 1 \\ \cline{2-10} 
\multicolumn{1}{|c|}{} & 6 & \multicolumn{1}{c|}{0.89} & \multicolumn{1}{c|}{0.93} & \multicolumn{1}{c|}{0.96} & 0.98 & \multicolumn{1}{c|}{0.37} & \multicolumn{1}{c|}{0.78} & \multicolumn{1}{c|}{0.59} & 0.97 \\ \cline{2-10} 
\multicolumn{1}{|c|}{\multirow{-4}{*}{3}} & 8 & \multicolumn{1}{c|}{0.97} & \multicolumn{1}{c|}{1} & \multicolumn{1}{c|}{0.98} & 1 & \multicolumn{1}{c|}{0.79} & \multicolumn{1}{c|}{0.96} & \multicolumn{1}{c|}{0.85} & 0.98 \\ \hline
\multicolumn{1}{|c|}{} & 10 & \multicolumn{1}{c|}{0.3} & \multicolumn{1}{c|}{0.38} & \multicolumn{1}{c|}{0.59} & 0.77 & \multicolumn{1}{c|}{0.07} & \multicolumn{1}{c|}{0.44} & \multicolumn{1}{c|}{0.32} & 0.83 \\ \cline{2-10} 
\multicolumn{1}{|c|}{} & 2 & \multicolumn{1}{c|}{0.79} & \multicolumn{1}{c|}{0.81} & \multicolumn{1}{c|}{0.87} & 0.92 & \multicolumn{1}{c|}{0.79} & \multicolumn{1}{c|}{0.86} & \multicolumn{1}{c|}{0.785} & 0.93 \\ \cline{2-10} 
\multicolumn{1}{|c|}{} & 4 & \multicolumn{1}{c|}{0.58} & \multicolumn{1}{c|}{0.72} & \multicolumn{1}{c|}{0.79} & 0.9 & \multicolumn{1}{c|}{0.82} & \multicolumn{1}{c|}{0.98} & \multicolumn{1}{c|}{0.75} & 0.95 \\ \cline{2-10} 
\multicolumn{1}{|c|}{\multirow{-4}{*}{9}} & 6 & \multicolumn{1}{c|}{-0.11} & \multicolumn{1}{c|}{-0.09} & \multicolumn{1}{c|}{0.54} & 0.59 & \multicolumn{1}{c|}{0.18} & \multicolumn{1}{c|}{0.57} & \multicolumn{1}{c|}{0.48} & 0.85 \\ \hline
\multicolumn{1}{|c|}{} & 8 & \multicolumn{1}{c|}{0.94} & \multicolumn{1}{c|}{0.97} & \multicolumn{1}{c|}{0.98} & 0.98 & \multicolumn{1}{c|}{0.93} & \multicolumn{1}{c|}{1} & \multicolumn{1}{c|}{0.95} & 1 \\ \cline{2-10} 
\multicolumn{1}{|c|}{} & 10 & \multicolumn{1}{c|}{0.47} & \multicolumn{1}{c|}{0.49} & \multicolumn{1}{c|}{0.74} & 0.76 & \multicolumn{1}{c|}{0.93} & \multicolumn{1}{c|}{1} & \multicolumn{1}{c|}{0.95} & 1 \\ \cline{2-10} 
\multicolumn{1}{|c|}{} & 2 & \multicolumn{1}{c|}{1} & \multicolumn{1}{c|}{1} & \multicolumn{1}{c|}{1} & 1 & \multicolumn{1}{c|}{0.85} & \multicolumn{1}{c|}{0.93} & \multicolumn{1}{c|}{0.91} & 0.97 \\ \cline{2-10} 
\multicolumn{1}{|c|}{\multirow{-4}{*}{7}} & 4 & \multicolumn{1}{c|}{1} & \multicolumn{1}{c|}{1} & \multicolumn{1}{c|}{1} & 1 & \multicolumn{1}{c|}{0.81} & \multicolumn{1}{c|}{0.91} & \multicolumn{1}{c|}{0.89} & 0.98 \\ \hline
\multicolumn{1}{|c|}{} & 8 & \multicolumn{1}{c|}{0.51} & \multicolumn{1}{c|}{0.63} & \multicolumn{1}{c|}{0.72} & 0.83 & \multicolumn{1}{c|}{0.15} & \multicolumn{1}{c|}{0} & \multicolumn{1}{c|}{0.49} & 93 \\ \cline{2-10} 
\multicolumn{1}{|c|}{} & 10 & \multicolumn{1}{c|}{0.87} & \multicolumn{1}{c|}{0.85} & \multicolumn{1}{c|}{0.95} & 0.95 & \multicolumn{1}{c|}{0.92} & \multicolumn{1}{c|}{1} & \multicolumn{1}{c|}{0.95} & 1 \\ \cline{2-10} 
\multicolumn{1}{|c|}{} & 2 & \multicolumn{1}{c|}{0.52} & \multicolumn{1}{c|}{0.55} & \multicolumn{1}{c|}{0.77} & 0.85 & \multicolumn{1}{c|}{0.46} & \multicolumn{1}{c|}{0.54} & \multicolumn{1}{c|}{0.49} & 0.8 \\ \cline{2-10} 
\multicolumn{1}{|c|}{\multirow{-4}{*}{5}} & 6 & \multicolumn{1}{c|}{0.39} & \multicolumn{1}{c|}{0.45} & \multicolumn{1}{c|}{0.65} & 0.75 & \multicolumn{1}{c|}{0.05} & \multicolumn{1}{c|}{0} & \multicolumn{1}{c|}{0.34} & 0.91 \\ \hline
\multicolumn{2}{|c|}{Average} & \multicolumn{1}{c|}{0.808} & \multicolumn{1}{c|}{0.8515} & \multicolumn{1}{c|}{0.623} & 0.6535 & \multicolumn{1}{c|}{0.7015} & \multicolumn{1}{c|}{0.9475} & \multicolumn{1}{c|}{0.5725} & 0.733 \\ \hline
\end{tabular}
\caption{IAA for bias and propaganda annotations.}
\label{tab:my:IAA}
\end{table*}
\subsection{Annotation Guidelines}
\label{sec:guidelines}
Our understanding of  "bias" is based on the work done by the United Nations Committee on the Elimination of Racial Discrimination and the European Commission against Racism and Intolerance \cite{EEAS}. We define the notations ‘bias’ and ‘propaganda’ based on the UN and EU accounts, as:

\textbf{Bias}: is generally understood as an inclination or prejudice towards or against a particular person or group, often in a way considered to be unfair. In other words, it is an unreasonable preference or dislike that prompts someone to behave in a discriminatory way, often based on unfair judgment. This bias is typically based on prohibited grounds of discrimination such as race, religion, language, nationality, ethnicity, social background, gender, and others.

\textbf{Classifications of Bias}: we adopted the same classes provided in the Shared Task: (1) Biased against Palestine,(2) Biased against Israel,
    (3) Biased against others,
    (4) Biased against both Israel and Palestine,
   (5) Not Applicable,
    (6) Unclear, and (7) Unbiased. We also introduced a new feature called "Type of Bias", which can be either: (\textit{a}) $Explicit$ (\Ar{تحيز صريح}) if it is obvious and evident in the post, (\textit{b}) $Implicit$ (\Ar{تحيز ضمني}) if it is clear but not evident in the post, and (\textit{c}) $Vague$ (\Ar{تحيز مبهم}) in case of indirect and ambiguous bias. This feature is important from a methodological viewpoint as it encourages the annotators to think more during classification. If a post contains biased content but not in a direct way it can be accounted as implicit.
\\
\textbf{Propaganda}: misleading ideas or statements that can distort the truth or omit facts to promote a specific political or social agenda. These ideas are typically published by media outlets.  For example, propaganda can take the forms of exaggeration, minimization, spreading doubts, name-calling, labeling, or intentional vagueness. All these forms have the common intention to spread false information and obscure facts.

\textbf{Classifications of Propaganda:} We adopted the four classes provided in the Shared Task: (i) \textit{Propaganda}, (ii) \textit{Not propaganda},
(iii) \textit{Not Applicable}, and 
(iv) \textit{Unclear}.

Additionally, we added a new column to classify Propaganda into three types: 
    (1) \textit{Propaganda must be deleted}: if it contains evident harmful content that poses risks to the safety and security of individuals or groups;
(2) \textit{Propaganda may be deleted}: if we cannot easily judge whether it is propaganda, depending on a specific context; and
(3) \textit{Propaganda not to be deleted}: if it is not clear and lacks harmful consequences and therefore does not warrant deletion.

\textbf{Remark:} Since the data was collected from Facebook posts some cases contain quoted content (e.g. an unbiased post quoting biased content). It was established in the guidelines that a post should not be classified as bias or propaganda based on its quotation, but rather on the post itself. 
 
\par \textbf{An Example} of the guidelines mentioned earlier regarding quoted content is as follows: “Hamas and Islamic Jihad spare no effort to exploit religious institutions for terrorist purposes,” the IDF said in a statement. This post is annotated as unbiased because it is a direct quote and does not include any additional commentary or interpretation.

\subsection{Inter-Annotator Agreement (IAA)}

To evaluate the quality of our annotations, we used the $F1$-score and Cohen's Kappa  \cite{cohen1968weighted} to compute the agreement between the annotators. The results are shown in Table \ref{tab:my:IAA}.

The task organizers allocated 100 posts ($10\%$) from each batch for IAA, including $20$ posts randomly selected from each language. Overall, we annotated $12,000$ posts, resulting in an IAA dataset of $1,200$ posts.  These were distributed among our $10$ annotators following this scheme: (1) each annotator received $240$ posts, (2) each post was annotated by two different annotators, and (3) the $240$ posts assigned to each annotator were distributed among four other annotators. Consequently, each pair of annotators had $60$ posts in common.

\textbf{All vs. Binary IAA}: 
to evaluate whether a (dis)agreement was dominated by a certain class, we mapped all labels into binary categories: ($Bias$ or $NotBias$ $and$ $others$) and ($Propaganda$ or $NotPropaganda$ $and$ $others$). Table \ref{tab:my:IAA} demonstrates no class dominance because All and Binary evaluations are close to each other.

Looking at all Cohen's scores in Table \ref{tab:my:IAA}, the average is $0.808$ for bias, which is a "very good" agreement, and  $0.7015$ for propaganda, which is a "good" agreement overall. Agreement on propaganda was more challenging but the results are enhanced when it is considered as a binary.

\section{Team Composition and Training}
\label{sec_team}
 \textbf{Team composition:} We assembled a team of $10$ Master’s students specializing in Law at Birzeit University, comprising $7$ females and $3$ males. All team members are native Arabic speakers with a good command of English. 

\textbf{Training phase}: We began by selecting $200$ posts to train all students in annotation. After training, each student was assigned $1,200$ posts for annotation.
\par \textbf{Ensuring consistency}
We held three workshops to ensure consistency to discuss guidelines, address challenges, and resolve disparities.The first workshop involved an expert who reviewed the annotations and added comments for the annotators to address. In the second workshop, the annotators met with the expert to discuss his comments on the posts. In the final workshop, after reviewing their annotations compared to the expert's, they discussed the points of agreement and disagreement with him.
\label{sec3}
\begin{table*}[t]
\small
\centering
\begin{tblr}{
  width = \linewidth,
  colspec = {Q[125]Q[119]Q[179]Q[163]Q[337]},
  cells = {c},
  hlines,
  vlines,
}
Subtask             & \textbf{Track} & \textbf{1st Place} & \textbf{2nd Place} & \textbf{3rd Place}                  \\
\textbf{Bias}       & Guidelines     & NLPColab           & Eagles             & Narrative Navigators                \\
\textbf{Bias}       & IAA Quality    & NLPColab           & JusticeLeague      & \textbf{Sina}                       \\
\textbf{Bias}       & Quantity       & DRAGON             & NLPColab           & \textbf{Sina}                       \\
\textbf{Bias}       & Consistency    & The Lexicon Ladies & NLPColab           & Narrative Navigators                \\
\textbf{Propaganda} & Guidelines     & NLPColab           & Bias Bluff Busters & \textbf{Sina}                                \\
\textbf{Propaganda} & IAA Quality    & NLPColab           & \textbf{Sina}      & The CyberEquity Lab                 \\
\textbf{Propaganda} & Quantity       & NLPColab           & \textbf{Sina}      & The CyberEquity Lab                 \\
\textbf{Propaganda} & Consistency    & NLPColab           & Bias Bluff Busters & Sahara Pioneers/The CyberEquity Lab 
\end{tblr}
\caption{FIGNEWS 2024 shared task results.} \label{result}
\end{table*}
\subsection{Annotation process}

 \textbf{Annotation Phase:} The dataset  consisted of $12$ batches, comprising $10,800$ posts from the \texttt{Main} sheet, and $1200$ posts from the \texttt{IAA} sheet. The annotation was carried out in two phases: 
\begin{enumerate}
    \item \textbf{Phase One}: We distributed \texttt{Batch01} and \texttt{Batch02}, each with $180$ posts, among team members. To ensure consistency with the guidelines, an expert reviewed all student annotations for these batches and provided feedback.
    \item  \textbf{Phase Two}: we assigned each annotator $450$ posts from two different batches. This step allowed us to complete the annotation of all $12$ batches (i.e. $12k$ posts).
\par \textbf{Set quality standards}
\par To set quality standards among annotators, after the annotation process was complete, each pair of annotators who had annotated the same data held meetings to review the selected posts they disagreed on. They discussed their differences, and if they reached an agreement, they would change the label accordingly. If they could not agree, they kept the original label.
\end{enumerate}

\section{Task Participation and Results}

\subsection{Results}
\label{sec_stat}
Table \ref{result} displays the final results provided by the shared task organizers. Our Sina team achieved the third and second place in the IAA Quality and Quantity tracks for the Bias and Propaganda subtasks, respectively. In addition to third place in Propaganda Guidelines.

Table \ref{tab:bias} and Table \ref{biased} illustrate the distribution of the bias classes and types of bias across languages respectively. Table \ref{tab:bias} shows that about $27\%$ of the posts are biased against Palestine and $63\%$ of the posts are unbiased. Most of the bias against Palestine originated from French posts.
Table \ref{biased} gives more statistics about the types of bias. As shown in this table, most of the posts annotated as \textit{$Explicit$} bias are in Hebrew.
\par For propaganda results, Table \ref{tab:propaganda} illustrates the distributions of propaganda classes across languages, which shows that $31$\% of the posts ($3333$) are annotated as "Propaganda", and $66\%$ ($7084$) are "Not Propaganda". The majority of the propaganda originated from French posts. 
Table \ref{tab:propaganda1} illustrates the distribution of the type of propaganda classes among languages. As shown in the table posts that were classified as propaganda must be deleted were in French with 348 posts.

\begin{table}[!ht]
\centering
\small % Reduce font size
\setlength{\tabcolsep}{1pt} % Adjust column separation
\begin{tabular}{|p{2.8cm}|c|c|c|c|c|c|} % Adjust the width as needed

\hline
\textbf{Class} & \textbf{Ar} & \textbf{En} & \textbf{He} & \textbf{Fr} & \textbf{Hi} & \textbf{Total} \\

\hline

Biased Against Palestine & 466 & 514 & 595 & 807 & 534 & \textbf{2916} \\ \hline
Biased Against Israel & 94 & 79 & 23 & 19 & 70 & \textbf{285} \\ \hline
Biased against Both & 6 & 7 & 11 & 6 & 14 & \textbf{44} \\ \hline
Biased against others & 42 & 28 & 53 & 39 & 49 & \textbf{211} \\ \hline
Unbiased & 1371 & 1486 & 1369 & 1212 & 1386 & \textbf{6824} \\ \hline
Not applicable & 49 & 7 & 17 & 20 & 25 & \textbf{118} \\ \hline
Unclear & 132 & 39 & 92 & 57 & 82 & \textbf{402} \\
\hline
\textbf{Total} & \textbf{2160} & \textbf{2160} & \textbf{2160} & \textbf{2160} & \textbf{2160} & \textbf{10800} \\\hline
\end{tabular}
\caption{Distribution of bias classes across languages}
\label{tab:bias}
\end{table}

\begin{table}[ht]
 \small 
 \setlength{\tabcolsep}{4pt}
\begin{tabular}{|l|r|r|r|r|r|r|}
\hline
\textbf{Type of Bias} & \textbf{Ar} & \textbf{En} & \textbf{He} & \textbf{Fr} & \textbf{Hi} & \textbf{Total} \\ \hline
\textit{$Explicit$} (\Ar{تحيز صريح}) & 394 & 336 & \textbf{563} & 412 & 388 & \textbf{2093} \\ \hline
\textit{$Implicit$} (\Ar{تحيز ضمني}) & 199 & 217 & 265 & 236 & 269 & \textbf{1186} \\ \hline
\textit{$Vague$} (\Ar{تحيز مبهم}) & 36 & 37 & 59 & 52 & 27 & \textbf{211} \\ \hline

\end{tabular}
\caption{Types of Bias}\label{biased}
\end{table}

\begin{table}[!ht]
\centering
\small % Reduce font size
\setlength{\tabcolsep}{1pt} % Adjust column separation
\begin{tabular}{|p{2.8cm}|c|c|c|c|c|c|} % Adjust the width as needed
\hline
\textbf{Class} & \textbf{Ar} & \textbf{En} & \textbf{He} & \textbf{Fr} & \textbf{Hi} & \textbf{Total} \\
\hline

Propaganda & 524 & 679 & 648 & 809 & 673 & \textbf{3333} \\\hline
Not propaganda & 1484 & 1443 & 1447 & 1297 & 1413 & \textbf{7084} \\ \hline
Not applicable & 48 & 11 & 17 & 16 & 24 & \textbf{116} \\\hline
Unclear & 104 & 27 & 48 & 38 & 50 & \textbf{267} \\
\hline
\textbf{Total} & \textbf{2160} & \textbf{2160} & \textbf{2160} & \textbf{2160} & \textbf{2160} & \textbf{10800} \\
\hline
\end{tabular}
\caption{Distribution of propaganda subtask classes across languages}
\label{tab:propaganda}
\end{table}

\begin{table}[!ht]
\centering
\small % Reduce font size
\setlength{\tabcolsep}{3.5pt} % Adjust column separation
\begin{tabular}{|l|c|c|c|c|c|c|} % Adjust the width as needed
\hline
\textbf{Class} & \textbf{Ar} & \textbf{En} & \textbf{He} & \textbf{Fr} & \textbf{Hi} & \textbf{Total} \\
\hline
\makecell{Propaganda Must \\ be deleted }& 192 & 191 & 266 &  \textbf{348} & 277 & \textbf{1274} \\ \hline
\makecell{Propaganda  May \\ be deleted}& 524 & 488 & 382 & 461 & 396 & \textbf{2059} \\\hline

\makecell{Propaganda not \\ to be deleted }& 451 & 422 & 565 & 648 & 436 & \textbf{2522} \\\hline
\end{tabular}
\caption{Types of Propaganda classes.}
\label{tab:propaganda1}
\end{table}

\section{Error Analysis and Discussion}
\label{sec_discussion}
Despite training and supervision, errors may arise from subjective interpretation, ambiguous guidelines, or complex content. We explored the errors and noted:
\begin{enumerate}
    \item False positives in bias annotations occurred when annotators marked neutral content as biased. For instance, the post: "Israel launched attacks on Syria on Nov $10$ in response to a drone strike on Eilat. The IDF claimed it attacked an organization responsible for the drone. Watch for more details." This news excerpt is informative and not biased.
    \item  Misclassification of propaganda: Some content was wrongly labeled as "must be deleted" propaganda despite lacking direct harmful implications. For example: "BREAKING: Israeli forces are causing massive destruction in Gaza, in response to a terrorist attack by Hamas. Image source: Middle East Eye post." While it is propaganda, it shouldn’t be classified as "must be deleted."
\end{enumerate}
 
%2. False positives in propaganda annotations: There were instances where neutral content was mistakenly labeled as propaganda. For example, in this post: "PUBG Mobile explodes a surprise with exciting details that excite players.", a promotional message about PUBG Mobile does not qualify as propaganda.

%\section{Compression with other teams}

%this section will be written after announcing the results of the shared task.  

\section{Conclusion}
\label{sec_conclusion}
This article presents our contribution to the FigNews $2024$, where we annotated a multilingual corpus of $12,000$ Facebook posts for bias and propaganda across five languages. We extended the annotation guidelines for better consistency and accuracy, providing a foundation for future work in detecting bias in social media. 
Our plans include expanding the corpus to cover more critical events of the war and leveraging neural and large Language models to automatically detect bias and propaganda on social media posts.

\section*{Ethical Considerations}
\label{sec_ethics}
Given the sensitive nature of the topics and media narratives related to the Israel War on Gaza, our annotators, who are lawyers, have undergone extensive training to ensure careful and fair judgments. They meticulously review both Arabic and English translations to avoid any bias that might arise from machine translation. 

\section*{Limitations}
\label{sec_limitations}
We recognize the limitations in our annotation process. This is because of the subjective nature of identifying bias and propaganda in social media posts, and the sensitivity of the datasets involved.

\section*{Acknowledgments}
We would like to acknowledge the contributions of Nejira Softic during the formulation of the guidelines. 
We would like to also thank the Master’s students specializing in Law and IT at Birzeit University for their help in annotating the datasets, especially Maram Shour, Belal Abu Zaina, Bayan Abu Alawi, Zainah Abughosh, Aya Al Dimasy, Doaa Abozena, Waad Alsheikh, Qassam Abu Hakmeh, Aseel Mustafa, Basel Awwad, Omar To’Mallah, and Dyala Fakhouri, and as well as Prof. Reem Al-Botmeh for her support during the course. Also thanks to Palestine Technical University - Kadoorie for its support.

\bibliography{custom,MyReferences}

%\printbibliography
\end{document}